\documentclass[conference]{IEEEtran}
\usepackage{times}

\usepackage[numbers]{natbib}
\usepackage{multicol}
\usepackage[bookmarks=true]{hyperref}
\usepackage{graphicx}
\usepackage{amsmath,amsfonts,amssymb}
\usepackage{xcolor}
\usepackage{lipsum}
\usepackage{booktabs}
\usepackage{multirow}
\usepackage{siunitx}
\usepackage{hyperref}

\hypersetup{
    pdftitle={Continuum Robot Localization using Distributed Time-of-Flight Sensors},
    pdfauthor={Spencer Teetaert},
    pdfsubject={Robots},
    pdfkeywords={State estimation,Soft robots}
}

\newcommand{\edit}[1]{{#1}}

\author{
    \IEEEauthorblockN{Spencer Teetaert\IEEEauthorrefmark{1}, Giammarco Caroleo\IEEEauthorrefmark{2}, Marco Pontin\IEEEauthorrefmark{2}, Sven Lilge\IEEEauthorrefmark{3},}
    \IEEEauthorblockN{Jessica Burgner-Kahrs\IEEEauthorrefmark{1}, Timothy D. Barfoot\IEEEauthorrefmark{1}, Perla Maiolino\IEEEauthorrefmark{2}}
    \IEEEauthorblockA{\IEEEauthorrefmark{1}University of Toronto Robotics Institute, Canada}
    \IEEEauthorblockA{\IEEEauthorrefmark{2}Oxford Robotics Institute, University of Oxford, UK}
    \IEEEauthorblockA{\IEEEauthorrefmark{3}Toronto Metropolitan University, Canada}
}

\begin{document}

% paper title
\title{Continuum Robot Localization using Distributed Time-of-Flight Sensors}

% You will get a Paper-ID when submitting a pdf file to the conference system

\maketitle

\begin{abstract}
Localization and mapping of an environment are crucial tasks for any robot operating in unstructured environments. Time-of-flight (ToF) sensors (e.g.,~lidar) have proven useful in mobile robotics, where high-resolution sensors can be used for simultaneous localization and mapping. In soft and continuum robotics, however, these high-resolution sensors are too large for practical use. This, combined with the deformable nature of such robots, has resulted in continuum robot (CR) localization and mapping in unstructured environments being a largely untouched area. In this work, we present a localization technique for CRs that relies on small, low-resolution ToF sensors distributed along the length of the robot. By fusing measurement information with a robot shape prior, we show that accurate localization is possible despite each sensor experiencing frequent degenerate scenarios. We achieve an average localization error of \SI{2.5}{cm} in position and \ang{7.2} in rotation across all experimental conditions with a \SI{53}{cm} long robot. We demonstrate that the results are repeated across multiple environments, in both simulation and real-world experiments, and study robustness in the estimation to deviations in the prior map. 
\end{abstract}

\IEEEpeerreviewmaketitle

\section{Introduction}
Continuum robots (CRs) are well-suited for many inspection and manipulation tasks in constrained environments due to their inherent compliance and capacity to be miniaturized~\cite{Russo2023}. This has led them to be studied for applications such as surgery~\cite{BurgnerKahrs2015, Dupont2022}, aircraft inspection~\cite{Chang2025,Dong2017,Wang2020}, and search and rescue operations~\cite{Yamauchi2022}. This miniaturization comes at the cost of additional size constraints on the sensors that can be integrated into the robot, making localization using CRs particularly challenging. Time-of-flight (ToF) sensors have been studied for CRs recently~\cite{Abah2022,Caroleo2025b,Iwao2025} though the sensor resolution, range, and field of view are limited at the smaller scales of CRs. This can lead to degenerate measurements where single ToF sensors are unable to observe enough features in the environment to fully constrain their positions. 

\begin{figure}[t]
    \centering
    \includegraphics[width=\linewidth]{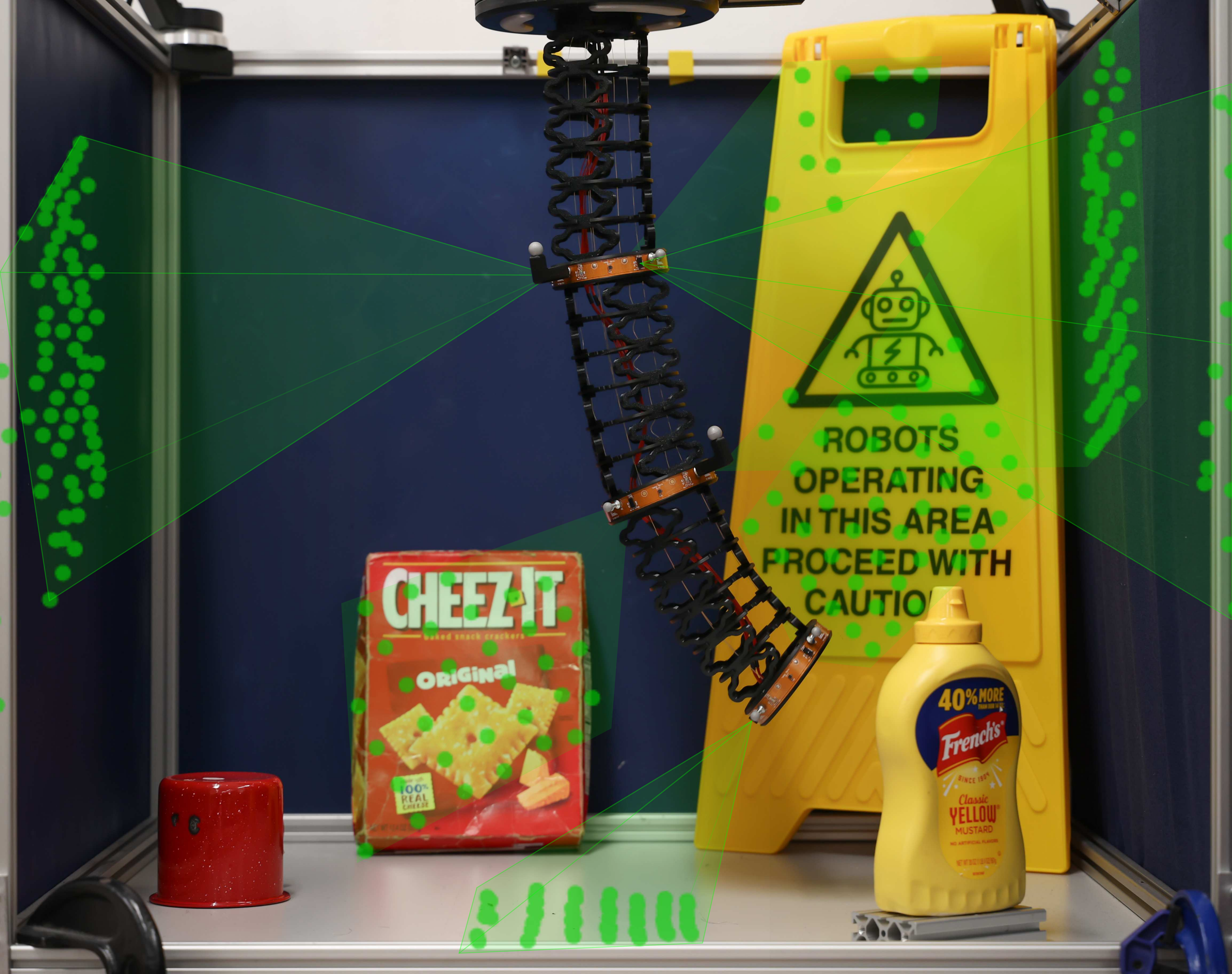}
    \caption{A continuum robot equipped with distributed sensors is shown localizing in a cluttered environment. The \SI{53}{cm} long, \SI{7.6}{cm} diameter extensible soft, continuum robot is equipped with three distributed sensor rings, each with three low-resolution ToF sensors and one gyroscope. Using only these sparse measurements, the robot is able to localize itself with high accuracy (\SI{1.9}{cm} average distance error in the shown environment). Noisy ToF sensor measurements are shown as green points alongside the ToF sensor fields of view. At many points in time, individual sensor rings do not see enough features to fully constrain their location, yet by fusing the distributed measurements with a robot shape prior, accurate localization is still possible.}
    \label{fig:title_fig}
\end{figure}

In previous works, distributed sensing has been shown to be an effective way to both localize and perform shape estimation for robots with an estimated configuration~\cite{Iwao2025,Sorensen2021,Teetaert2025a,Yamauchi2022}. By providing multiple sensing locations along the body of the robot, distributed sensing can provide a richer understanding of both the robot's configuration and its surrounding environment. Additionally, recent works have demonstrated the effectiveness of continuous-time state estimation for CRs~\cite{Teetaert2025a,Teetaert2025b}, allowing for state estimation from both low- and high-frequency, asynchronous sensor measurements. This framework is particularly well-suited for CRs as it enables the fusion of measurements from multiple distributed sensors along the robot's body with varying sampling rates. 

Despite recent progress in distributed sensing and continuous-time state estimation for continuum robots, most existing localization approaches rely on either dense sensing, rigid-body assumptions, or external reference systems. These assumptions break down for small-diameter, deformable robots operating in cluttered, unstructured environments, where onboard sensors are necessarily sparse, low-resolution, and frequently produce geometrically degenerate measurements. In such settings, individual sensors are insufficient to fully constrain robot pose, and even moderate deviations between the prior map and the true environment can lead to estimator failure. In this work, we address this gap by demonstrating that full-body, onboard-only localization of a deformable continuum robot is feasible using sparse, distributed time-of-flight sensing, even in the presence of frequent measurement degeneracies and partial map mismatch. The contributions of this work are: 
\begin{itemize}
    \item A continuous-time, factor-graph-based MAP estimation framework that integrates sparse time-of-flight measurements with a robot shape prior, enabling stable full-body localization of deformable continuum robots under frequent geometric degeneracies and partial prior map mismatch.
    \item The first real-world demonstration of onboard-only localization of a soft, extensible continuum robot in cluttered environments using distributed, low-resolution ToF and inertial sensing, validated across both simulated and physical platforms.
\end{itemize} The code used in this work is available to the community at \url{https://github.com/utiasASRL/space_time_continuum}.

\section{Related Work}
% Application and problem motivation
Continuum and snake-like robots have been proposed for a number of inspection and maintenance tasks in constrained environments. Their ability to be miniaturized~\cite{Zhang2022} enables applications in confined and convoluted environments. Both \cite{Chang2025} and~\cite{Wang2020} explore the use of CRs for aircraft inspection, while~\cite{Dong2017} demonstrates basic machining tasks in a similar environment. This application space has even seen some commercial success with articulated borescopes~\cite{GEAero}. While many previous works have focused on shape estimation of CRs~\cite{Ferguson2024, Lilge2022, Rone2013, Shi2017}, we focus rather on the task of localizing the robot with respect to some inertial reference frame. To enable any degree of autonomy in the aforementioned applications, accurate localization with respect to an environment is critical. 

% Off-body localization 
Localization of CRs has been explored through a variety of sensing modalities, including cameras, ToF sensors, and electromagnetic (EM) trackers~\cite{Sincak2024}. Many existing works use sensors with off-robot components for localization, including ultrasound~\cite{Koolwal2010}, electromagnetic pose sensors~\cite{Franz2014}, fluoroscopic-like stereo imaging~\cite{Borgstadt2015}, and external cameras~\cite{Reiter2011, Shentu2023}. These approaches often require line-of-sight to the robot or specialized imaging equipment to sense the robot's state, which may not be available in many applications. One alternative is to mount the sensors directly on the robot itself and use them to sense the robot's environment, inferring location from relative positioning. Such approaches are typically more challenging due to the size and weight constraints of the sensors. 

% Camera localization
Some works have shown promise with camera-based localization~\cite{Girerd2020, Sorensen2021, Yamauchi2022}. In~\cite{Vaquero2024}, cameras are used alongside a ToF sensor (lidar) for localization of a snake-like robot. A tip camera as well as a fiber Bragg grating (FBG) sensor is used in~\cite{Wang2023} to estimate robot pose. In~\cite{Sorensen2021}, multiple cameras are distributed along a hybrid soft-rigid robot to perform visual SLAM. Through sensor distribution, the authors are able to achieve full-state localization and configuration estimation. As for sensor distribution, \cite{Mahoney2016} proposes an uncertainty-based framework for optimizing sensor placement along CRs while~\cite{Kim2014} explores optimal placement of fiber Bragg gratings for shape sensing. Both of these works highlight the importance of sensor placement for accurate state estimation beyond just at the robot tip, though the latter only measures intrinsic information about the robot's shape and not the location of the robot within a broader environment.

% ToF sensors 
ToF sensing offers different advantages to cameras, enabling native depth perception and increased robustness in visually degraded environments. At smaller scales, however, dense ToF measurements are typically not available. Some existing works have explored the use of sparse ToF sensors in CR applications. In~\cite{Iwao2025}, a series of 40-160 small ToF sensors are used for full-body localization of a rigid snake robot. The authors solve for spatial states in series along the robot body in a simulated environment. In~\cite{Caroleo2025a}, a characterization of the VL53L5CX ToF sensor is provided, while~\cite{Caroleo2025b} proposes a state estimation algorithm using the sensor on a soft robot. In~\cite{Abah2022}, multiple single-beam ToF sensors (VL6180X) are integrated into a proposed sensor module for CRs. The authors highlight the concept of moving the robot to create a denser map of the environment, but require external motion capture for localization. Our method takes a step towards the realization of the vision of~\cite{Abah2022} by introducing an algorithm for localizing a CR using sparse onboard ToF measurements without the need for external motion capture, marking the first demonstration of this capability on a soft continuum robot. We additionally move beyond the rigid-body assumption of~\cite{Iwao2025} to consider the challenges of localizing a deformable continuum robot.
\section{Estimation Framework}

The estimation framework used in this work is based on the continuous-time factor-graph method presented in~\cite{Teetaert2025a, Teetaert2025b}. The approach presents a factor-graph-based method to state estimation for CRs through a maximum a posteriori (MAP) optimization scheme. Prior factors are introduced that model a CR, with measurement factors being added for each sensor type. Factors are constructed using a weighted squared-error cost function. The total cost is minimized using a Gauss-Newton solver with uncertainties estimated via a Laplace approximation. 

We use the same sliding-window formulation and prior factors, expanding upon the system with the introduction of new measurement factors described in the following sections. Our state $\boldsymbol{x}(s, t)$ consists of the continuous set of robot poses, $\boldsymbol{T}_{ib}(s, t) \in SE(3)$, body-centric generalized strains, $\boldsymbol{\epsilon}(s, t) \in \mathbb{R}^6$, and body-centric generalized velocities, $\boldsymbol{\varpi}(s, t) \in \mathbb{R}^6$ at time $t$ and arc  length $s$. $\boldsymbol{T}_{ib}(s, t)$ represents an $SE(3)$ transformation matrix that transforms points from a robot body frame, $b$, to an inertial frame, $i$. During optimization, the state is estimated at discrete times, $t_k$, and arc lengths, $s_n$, with the full continuous representation being available through the interpolation scheme described in~\cite{Teetaert2025a}.

\subsection{Time-of-Flight Sensor}\label{sec:tof_factor}

To include the ToF measurements into the factor-graph estimation framework, we follow the work of~\cite{Burnett2025} and introduce a point-to-plane measurement factor with a Cauchy loss. Specifically, given a set of measured lidar distances $d_j \in \mathbb{R},\; j = 0,1,\hdots,M$ from a sensor on the robot at time $t$ and arc  length $s$, we define a point-to-plane error as
\begin{equation}
    \boldsymbol{e}_j(\boldsymbol{x}) = \alpha \boldsymbol{n}_j^T \boldsymbol{D}(\boldsymbol{p}_j - \boldsymbol{T}_{ib}(t, s)\boldsymbol{q}^b_j)
\end{equation}
where $\boldsymbol{p}_j$ is a matched point from a local map, $\boldsymbol{n}_j$ is the estimated normal at point $\boldsymbol{p}_j$, $\boldsymbol{D} \in \mathbb{R}^{3 \times 4}$ is a projection matrix that removes the homogeneous component, and $\alpha = (\sigma_2 - \sigma_3) / \sigma_1$ is a heuristic weight to favor planar neighborhoods. $\sigma_i$ is the square root of the $i$th eigenvalue of the covariance obtained through principal component analysis of the neighborhood. See~\cite{Demantke2011} for more details. $\boldsymbol{q}^b_j$ is a query point corresponding to measurement $j$ and is assumed to be corrupted by isotropic Gaussian noise,
\begin{align}
    \boldsymbol{q}^b_j &= \bar{\boldsymbol{q}}^b_j + \delta \boldsymbol{q}^b_j & \delta \boldsymbol{q}^b_j &\sim \mathcal{N}(\boldsymbol{0}, \boldsymbol{D}^T\boldsymbol{R}_{\text{ToF}}\boldsymbol{D}).
\end{align}Here, $\boldsymbol{R}_{\text{ToF}}$ is tuned experimentally starting with the relationships described in Table~\ref{tab:tof_covariance} with some additional noise to account for unmodeled process errors.

\begin{table}[t]
\centering
\caption{ToF distance-based standard deviation model.}
\begin{tabular}{c|c}
\toprule
\textbf{Distance measurement [m]} & \textbf{Standard deviation [m]} \\
\midrule
    $d < 0.025$ & Measurement invalid \\
    $0.025 \leq d < 0.6$ & $0.014d \rightarrow 0.012d$ \\
    $0.6 \leq d < 1.2$ & $0.012d \rightarrow 0.006d$ \\
    $1.2 \leq d$ & $0.006d$ \\
\bottomrule
\end{tabular}\label{tab:tof_covariance}
\vspace{0.5em}

\begin{minipage}{\linewidth}
    \raggedright
    \footnotesize
    $A \rightarrow B$ indicates a linear interpolation from value $A$ to value $B$ over the specified distance range.
\end{minipage}
\end{table}

As ToF sensors are subject to disturbances that can lead to outlier measurements, we use a robust cost function to mitigate their effect on the estimation. As the estimation framework employed in this work uses a Gauss-Newton scheme designed for squared error factors, we implement a Cauchy loss via an iteratively reweighted least squares (IRLS) scheme~\cite{Barfoot2017, Holland1977}. The final cost function for a single point-to-plane measurement is given by 
\begin{align}
    \nonumber J_j(\boldsymbol{x}) &= \frac{1}{2} \boldsymbol{e}_j(\boldsymbol{x})^T \boldsymbol{Y}_j(\boldsymbol{x})^{-1} \boldsymbol{e}_j(\boldsymbol{x}), \\
    \boldsymbol{Y}_j(\boldsymbol{x})^{-1} &= \frac{\boldsymbol{R}_{\text{ToF}}^{-1}}{1 + \boldsymbol{e}_j(\boldsymbol{x})^T \boldsymbol{R}_{\text{ToF}}^{-1} \boldsymbol{e}_j(\boldsymbol{x})}.
\end{align}
As with other factors in the framework, we must compute the linearized error for the Gauss-Newton update step. This is given by
\begin{align}
    \nonumber \boldsymbol{e}_j(\boldsymbol{x}) &= \bar{\boldsymbol{e}}_j(\boldsymbol{x}) + \left. \boldsymbol{G}_j\right|_{\boldsymbol{x}(s, t)}\delta \boldsymbol{t}_{bi}, \\
    \left. \boldsymbol{G}_j\right|_{\boldsymbol{x}(s, t)} &= \alpha \boldsymbol{n}_j^T \boldsymbol{D}(\bar{\boldsymbol{T}}_{ib}(t, s)\boldsymbol{q}^b_j)^\odot \bar{\boldsymbol{\mathcal{T}}}_{ib}(s, t).
\end{align} This linearization is performed in an $SE(3)$-sensitive way according to the methods described in~\cite{Barfoot2017}. $\delta \boldsymbol{t}_{bi} \in \mathbb{R}^6$ is the perturbation vector for the robot pose at time $t$ and arc  length $s$, $\boldsymbol{\mathcal{T}}_{ib}(s, t)$ is the adjoint matrix of the robot pose, $\text{Ad}(\boldsymbol{T}_{ib}(s, t))$, and $(\cdot)^\odot$ is the operator defined as 
\begin{align}
    \begin{bmatrix}
        \boldsymbol{a} \\ b
    \end{bmatrix}^\odot &= \begin{bmatrix}
        b\boldsymbol{I}_{3\times3} & -\boldsymbol{a}^\wedge \\
        \boldsymbol{0}_{1\times3} & 0
    \end{bmatrix}, \\
    \boldsymbol{a}^\wedge &= \begin{bmatrix}
        0 & -a_3 & a_2 \\
        a_3 & 0 & -a_1 \\
        -a_2 & a_1 & 0
    \end{bmatrix}.
\end{align} 

\subsection{Gyroscope Sensor}
In addition to the ToF sensors, gyroscope measurements are also included in the factor graph to aid in localization. The gyroscope measurements are modeled as in~\cite{Teetaert2025a}. As we are working on small time scales, gyroscope biases are approximated to be constant and are determined by averaging an initial stationary bias. 

\subsection{Strain Sensor (Simulation Only)}
The strain data used in this work has a bending angle and curvature representation that is consistent with many FBG sensing systems, enabling future hardware integration in more complex settings. Bending angle $\theta_j$ and curvature $\kappa_j$ are mapped to a measured strain value $\tilde{\boldsymbol{\epsilon}}_j$ via 
\begin{align}
    \tilde{\boldsymbol{\epsilon}}_j &= \boldsymbol{\mathcal{T}}(\theta_j) \begin{bmatrix}
        1 & 0 & 0 & 0 & 0 & \kappa_j
    \end{bmatrix}^T, \\
    \boldsymbol{\mathcal{T}}(\theta_j) &= \text{Ad} \left(\begin{bmatrix}
        1 & 0 & 0 & 0 \\
        0 & \cos(\theta_j) & -\sin(\theta_j) & 0 \\
        0 & \sin(\theta_j) & \cos(\theta_j) & 0 \\
        0 & 0 & 0 & 1
    \end{bmatrix}\right).
\end{align}
An associated error factor is used via the cost function 
\begin{equation}
    J_{j} = \frac{1}{2}(\boldsymbol{\epsilon}_j - \tilde{\boldsymbol{\epsilon}}_j)^T \boldsymbol{R}_{\text{strain}}^{-1} (\boldsymbol{\epsilon}_j - \tilde{\boldsymbol{\epsilon}}_j),
\end{equation} where $\boldsymbol{R}_{\text{strain}}$ is the strain measurement covariance and is modeled to be isotropic.

\subsection{Prior Map}\label{sec:prior_map_generation}
Each localization instance begins with a prior map of the environment represented as a point cloud with precomputed normals and planarity weights. For simulated environments, this map is generated via Poisson sampling of a CAD-generated scene. For real-world scenes, a point cloud is generated via scanning the scene with a Creality Raptor 3D scanner~\cite{3d_scanner}. Reflective markers are distributed in the scene for coordinate alignment to a motion capture system, and the near-infrared scanning mode is adopted, achieving a \SI{0.1}{mm} accuracy with a \qtyrange{0.1}{2}{mm} resolution. The map is stored in a hashed voxel map to enable fast nearest neighbor queries~\cite{Burnett2025}, though other representations (e.g.,~KD trees, PH-trees) are compatible with the work presented here. By varying which prior map the estimator has access to while keeping the sensor data the same, we can create scenarios where differences between the prior map and the scene are introduced. This is how deviations to the environment are created in both simulation and real-world experiments. 

\subsection{Scene Reconstruction}\label{sec:scene_reconstruction}
After estimation is complete, the robot's trajectory and ToF measurements are used to reconstruct a point cloud of the environment. Each ToF measurement can be projected into the inertial frame using the estimated robot pose at the time of the measurement using the uncertainty-aware projection,
\begin{align}
    \bar{\boldsymbol{p}}_j &= \bar{\boldsymbol{T}}_{ib}(t, s)\bar{\boldsymbol{q}}^b_j, \\
    \nonumber \delta \boldsymbol{p}_j &= \begin{bmatrix} -\bar{\boldsymbol{p}}_j^\odot \bar{\boldsymbol{\mathcal{T}}}_{ib} & \bar{\boldsymbol{T}}_{ib}  \end{bmatrix} \begin{bmatrix}
        \delta \boldsymbol{t}_{bi} \\ \delta \boldsymbol{q}^b_j
    \end{bmatrix} \sim \mathcal{N}(\boldsymbol{0}, \boldsymbol{D}^T\boldsymbol{\Sigma}_{p_j}\boldsymbol{D}), \\ 
    \label{eq:projected_covariance}\boldsymbol{\Sigma}_{p_j} &= \boldsymbol{R}_{\text{ToF}} + \boldsymbol{D}\bar{\boldsymbol{p}}_j^\odot \bar{\boldsymbol{\mathcal{T}}}_{ib} \boldsymbol{\Sigma}_{b} \bar{\boldsymbol{\mathcal{T}}}_{ib}^T \bar{\boldsymbol{p}}_j^{\odot T}\boldsymbol{D}^T.
\end{align} Here, $\boldsymbol{\Sigma}_{b}$ is the body-frame pose covariance as estimated during the optimization process. The expression in~\eqref{eq:projected_covariance} makes use of the fact that $\boldsymbol{R}_{\text{ToF}}$ is isotropic. 

\section{Simulation}
For generating simulated data, we use the MuJoCo physics engine~\cite{Todorov2012} with a number of custom additions to simulate a CR operating in a confined environment with ToF, gyroscope, and strain sensors. \edit{The simulated scene is designed to emulate the real-world application of jet engine inspection. The scene dimensions, robot specs, and anomaly choices are meant to reflect this.}

\subsection{Robot Details}
The simulated robot is modeled in MuJoCo as a series of rigid links connected by spherical joints. This approximation follows the pseudo-rigid body method commonly used in CR literature~\cite{Rao2021}. Provided the links are short enough, this robot will approximate the behavior of a true CR, while allowing it to take advantage of MuJoCo's extensive contact and dynamics modeling capabilities. The robot \edit{has three tendon-actuated modules with a combined length of \SI{1.5}{m}}, with each MuJoCo link being \SI{1}{cm} long. Every \SI{30}{cm} is a sensor disc containing three radially mounted ToF sensors and one gyroscope sensor. The discs are offset by one another by an axial rotation of \ang{60}. The tip of the robot contains an additional ToF sensor pointing forward. A strain sensor is simulated down the center of the robot, providing bending strain measurements every \SI{3}{cm}. See section~\ref{sec:sensors} for more information.

\subsection{Environment Setup}
\begin{figure}[t]
    \centering
    \includegraphics[width=\linewidth]{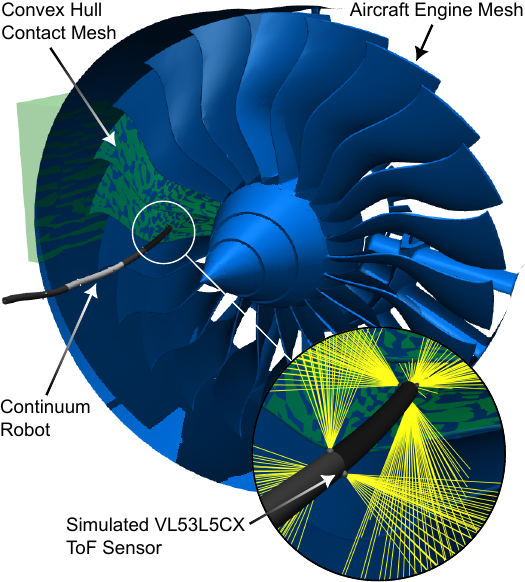}
    \caption{Simulated jet engine environment with CR in the MuJoCo physics engine. The robot is \SI{1.5}{m} long and contains distributed ToF, IMU, and strain sensors. The environment mesh is displayed in blue while the collision mesh is shown in green. Time-of-flight sensor rays are visualized in yellow.}
    \label{fig:simulation_setup}
\end{figure}
Motivated by the task of aircraft inspection~\cite{Chang2025,Dong2017,Wang2020}, the environment used in this work comes from a model jet engine~\cite{Catiav5ftw2025}. The model is scaled to have an outer diameter of \SI{3.1}{m} (approximately the size of a commercial airliner engine) and is cropped to contain only the front compressor section. The mesh is down-sampled to adhere to MuJoCo's \num{200,000} face limit on imported objects. A small section of the mesh where robot-environment collisions are expected is decomposed into a series of convex hulls using CoACD~\cite{Wei2022} to improve collision performance. Gravity, friction, and collisions are all enabled in the simulation. Fig.~\ref{fig:simulation_setup} shows the robot and environment used in the simulation.

Two types of map differences are added to the environment: unmodeled objects and missing modeled features. Unmodeled objects are simulated by adding additional meshes into the environment to be sensed and collided with by the robot. Missing modeled features are simulated by selectively removing parts of the environment mesh. In total, 10 environments are created: three with unmodeled objects (scenes S1--S3), three with missing modeled features (scenes S4--S6), three with both types of anomalies (scenes S7--S9), and one with no changes at all (scene S0). The anomalies range in size and are placed in locations that are reachable by the robot. 

\subsection{Sensor Simulation} \label{sec:sensors}
\subsubsection{Time-of-flight Sensor}
The ToF sensors are designed to simulate the VL53L5CX sensors from STMicro-Electronics used on the real robot in this work. It is simulated by a grid of $8\times 8$ MuJoCo rangefinder sensors with a shared origin and field of view of \ang{45} both horizontally and vertically. Each rangefinder has a maximum range of \SI{4}{m}. Noise is added to each range measurement according to the model described in Table~\ref{tab:tof_covariance}. Each ToF sensor generates data at a rate limited to at most \SI{15}{Hz}.

\subsubsection{Gyro Sensor}
The gyro sensors are modeled using the Mujoco gyroscope sensor. Noise is added to the gyroscope measurements using a zero-mean Gaussian with a standard deviation of \SI{0.01}{rad/s}. The sensor data is sampled at a rate of \SI{100}{Hz}.

\subsubsection{Strain Sensor}
Strain sensing is simulated through a custom MuJoCo sensor plug-in by computing the bending angle and curvature at specified points along the robot. To map from the pseudo-rigid body representation of the robot to a continuous strain profile a constant-curvature interpolation scheme is employed. Specifically, each link's midpoint is treated as the endpoint of a constant-curvature arc whose ends are tangent to the rigid link at the midpoint. This piecewise constant-curvature approximation enables strain estimation at any point along the robot. Zero-mean Gaussian noise with standard deviations of \SI{0.01}{m^{-1}} and \SI{0.015}{rad} is added to the curvature and bending-angle measurements, respectively. Strain data is sampled at a rate of \SI{20}{Hz}. 

\section{Experimental Setup}
\begin{figure}[t]
    \centering
    \includegraphics[width=\linewidth]{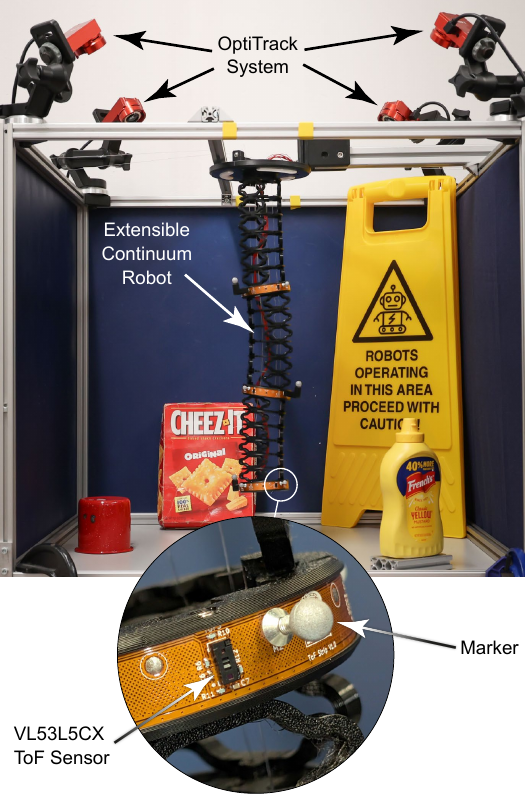}
    \caption{Experimental setup including the extensible continuum robot, the OptiTrack camera system, and a close-up of one of the sensorized rings carrying the markers, ToF sensors, and IMUs. The scene depicted contains all four objects present (scene R4).}
    \label{fig:experiment_setup}
\end{figure}
The experimental setup, shown in Fig.~\ref{fig:experiment_setup}, comprises a cube-shaped environment (\SI{70}{cm} $\times$ \SI{70}{cm} $\times$ \SI{60}{cm}  W$\,\times\,$D$\,\times\,$H), four OptiTrack cameras, a set of objects and a soft, continuum robot arm. The arm consists of three modules, each having three flexures printed out of Thermoplastic Polyurethane (TPU) 82A (Recreus) and a sensorized ring containing three ToF sensors and a gyroscope unit. Both the flexures and the ToFs for each module were oriented at \ang{120} from one another, and each module was offset by \ang{60} with respect to the one above it. At rest, the robot measures in at \SI{53}{cm} in length, with each module being approximately the same length. \edit{The experimental scene is designed to stress-test the limits of the proposed algorithm, with progressive anomalies increasing in size until localization failure is achieved. }

\subsection{Sensor Characteristics}
The ToF sensors used in this work are the VL53L5CX sensors from STMicro-Electronics. Each sensor has a horizontal and vertical field of view of 45° and produces an $8\times8$ grid of ToF measurements. The sensor noise characteristics of the VL53L5CX sensor were experimentally determined in~\cite{Caroleo2025a}, where it was found that the sensor noise can be modeled as a series of Gaussian distributions dependent on the measured range. Specifically, the measurement noise standard deviation for a single measurement is given in Table~\ref{tab:tof_covariance}. Each ring also incorporates an ISM330BX IMU (STMicro-Electronics) setup with an accelerometer full-scale of $\pm 4\,$g and a gyroscope full-scale range of $\pm 500\,$dps. For this work, only the gyroscope measurements are used in the estimation framework. The IMU is sampled at \SI{120}{Hz} while the ToF sensors are sampled at \SI{15}{Hz}.  

\subsection{Data Collection and Environments}\label{sec:data_collection}
A total of five environments were created for experimental validation, each with one more object added to the scene than the previous. The objects added in order are: a mustard bottle (scene R1), a cup (scene R2), a cracker box (scene R3), and a wet floor sign (scene R4), with the final environment being a scene with no objects (scene R0). The objects range in size and shape to provide a variety of conditions for testing. Each environment is scanned using the 3D scanner described in section~\ref{sec:prior_map_generation} to create a prior map for localization. A total of 23 trajectories are collected across the five environments, with all having at least four trajectories. During each trajectory, four tendons actuated the robot: one in the central module and three in the distal one. Due to the compliance of the flexures, the arm displays significant compression when being actuated, rendering any Kirchhoff rod approximations infeasible. 
\section{Results and Discussion}

For all results shown (real-world and simulation), the estimator is initially tuned using one hold-out trial consisting of a single trajectory operated in a fully known environment (R4 for real-world and S0 for simulation). Tuning is done by sweeping through reasonable values for the process and measurement noise parameters and selecting the set that minimizes the average localization error at the marker locations. The same tuned parameters are then used for all other trials. 

\subsection{Simulation}

% Simulation results 
\begin{table}[t]
\centering
\caption{Localization error in simulated environment using scene S0 as a prior map. Values shown as mean $\pm$ standard deviation across all trials and rings.}
\begin{tabular}{c|cc|cc}
\toprule
Scene &
\multicolumn{2}{c|}{Translation Error (cm)} &
\multicolumn{2}{c}{Rotation Error (deg)} \\
 & MAE & RMSE & MAE & RMSE \\
\midrule
S0  & 0.78 $\pm$ 0.48 & 0.99 $\pm$ 0.65 & 0.96 $\pm$ 0.48 & 0.98 $\pm$ 0.48 \\
\midrule
S1  & 0.78 $\pm$ 0.48 & 0.99 $\pm$ 0.65 & 0.96 $\pm$ 0.48 & 0.98 $\pm$ 0.48 \\
S2  & 0.78 $\pm$ 0.47 & 1.00 $\pm$ 0.64 & 0.96 $\pm$ 0.49 & 0.98 $\pm$ 0.48 \\
S3  & 0.78 $\pm$ 0.48 & 0.99 $\pm$ 0.65 & 0.96 $\pm$ 0.48 & 0.98 $\pm$ 0.48 \\
\midrule
S4  & 0.61 $\pm$ 0.34 & 0.74 $\pm$ 0.38 & 0.96 $\pm$ 0.46 & 0.98 $\pm$ 0.45 \\
S5  & 0.73 $\pm$ 0.43 & 0.94 $\pm$ 0.58 & 0.96 $\pm$ 0.48 & 0.98 $\pm$ 0.48 \\
S6  & 0.77 $\pm$ 0.47 & 0.98 $\pm$ 0.64 & 0.96 $\pm$ 0.48 & 0.98 $\pm$ 0.48 \\
\midrule
S7  & 0.61 $\pm$ 0.34 & 0.74 $\pm$ 0.38 & 0.96 $\pm$ 0.46 & 0.98 $\pm$ 0.45 \\
S8  & 0.76 $\pm$ 0.44 & 0.96 $\pm$ 0.60 & 0.96 $\pm$ 0.49 & 0.98 $\pm$ 0.48 \\
S9  & 0.77 $\pm$ 0.47 & 0.98 $\pm$ 0.64 & 0.96 $\pm$ 0.48 & 0.98 $\pm$ 0.48 \\
\bottomrule
\end{tabular}\label{tab:sim_results}
\end{table}

\begin{figure}[t]
    \centering
    \includegraphics[width=\linewidth]{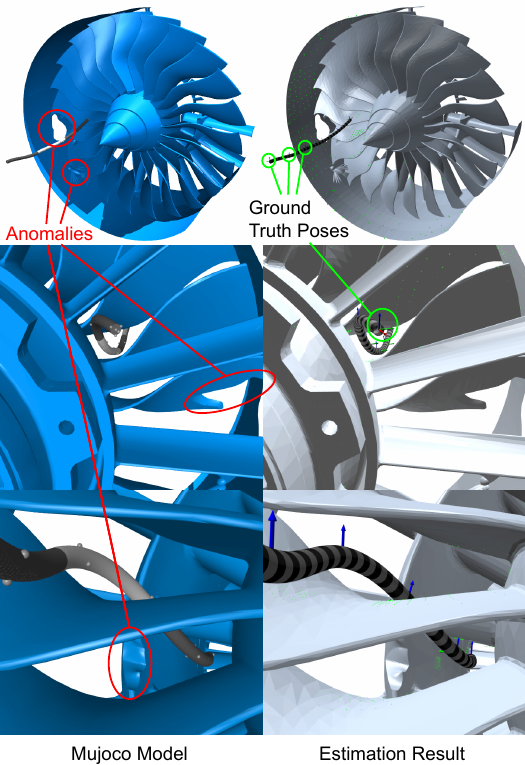}
    \caption{Side-by-side comparison of the simulated environment (left) and the estimated robot (right). The estimated robot shape and trajectory closely match the ground truth, with an average position error at the marker locations of less than \SI{1}{cm} and rotational error of less than \ang{1}. Anomalies in the environment are marked---they appear to have little effect on localization performance in simulation.}
    \label{fig:sim_side_by_side}
\end{figure}
\begin{figure}[t]
    \centering
    \includegraphics[width=\linewidth]{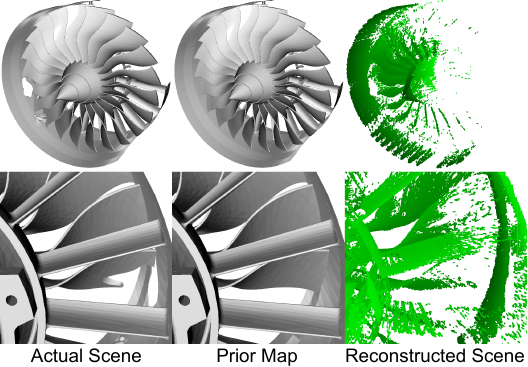}
    \caption{Simulated scene S8 (left), prior map S0 (middle), and the reconstructed scene (right) via ToF measurements. A depth-based color gradient is applied to the point clouds for visual clarity. The small objects added and removed between the two scenes appear to have little effect on localization performance in simulation.}
    \label{fig:sim_S8_pcd}
\end{figure}

To isolate the effect of prior map accuracy on localization performance, we conducted a series of simulation experiments where the robot actuation was kept identical while varying the scenes. Three robot trajectories are executed in each of the 10 scenes (S0--S9). For each of the 30 trials, the prior map used was that of scene S0. The results are summarized in Table~\ref{tab:sim_results}. The simulation results show no noticeable degradation in localization performance when the prior map deviates from the true environment due to the addition or removal of geometries from the map. This is made possible through the use of the robust cost function employed with the ToF factors in this work. The accuracy shown in the simulation deviates greatly from the real-world results, primarily due to the role of the strain sensor data in the estimate, as well as the idealized nature of the simulated environment and comparatively small map deviations, among other sources of unmodeled error. \edit{The strain sensor alone, in the absence of time-of-flight, does not provide enough information for accurate localization, resulting in `spatial drift' in the estimates. In this case, the error in the estimate increases from the base towards the tip, resulting in a large dip in tip accuracy compared to the fused time-of-flight and strain sensor setup. Likewise, time-of-flight without strain measurements results in estimates that frequently converge to local minima, due in large part to the radial symmetry of the environment. }

Side-by-side comparisons of the estimated robot shape against the simulated robot for the S8 scene are shown in Fig.~\ref{fig:sim_side_by_side}. Fig.~\ref{fig:sim_S8_pcd} shows a side-by-side of the true scene, prior map, and reconstructed scene via ToF measurements and the method described in Section~\ref{sec:scene_reconstruction}. The scene is constructed using all estimates collected on scene S8. In the simulated environment, the small anomalies present are not large enough to significantly affect localization and scene reconstruction accuracy. 

\subsection {Real World Experiments}

% Real world results
\begin{table}[t]
\centering
\caption{Localization error grouped by experiment condition.
Values shown as mean $\pm$ standard deviation across all trials and motion capture rings.}
\begin{tabular}{c|cc|cc}
\toprule
Map -- Scene &
\multicolumn{2}{c|}{Translation Error (cm)} &
\multicolumn{2}{c}{Rotation Error (deg)} \\
 & MAE & RMSE & MAE & RMSE \\
\midrule
\multicolumn{5}{c}{Localization within known environment (no anomalies)} \\
\midrule
R0 -- R0 & 1.8 $\pm$ 0.6 & 1.9 $\pm$ 0.7 & 5.2 $\pm$ 1.4 & 5.8 $\pm$ 1.8 \\
R1 -- R1 & 1.9 $\pm$ 0.9 & 2.3 $\pm$ 1.3 & 6.3 $\pm$ 4.3 & 7.8 $\pm$ 6.0 \\
R2 -- R2 & 2.0 $\pm$ 1.0 & 2.4 $\pm$ 1.3 & 6.3 $\pm$ 3.9 & 7.9 $\pm$ 5.5 \\
R3 -- R3 & 1.7 $\pm$ 0.8 & 1.9 $\pm$ 1.0 & 5.9 $\pm$ 2.6 & 6.8 $\pm$ 3.7 \\
R4 -- R4 & 1.9 $\pm$ 0.4 & 2.0 $\pm$ 0.4 & 8.4 $\pm$ 5.0 & 9.1 $\pm$ 5.1 \\
\midrule
\multicolumn{5}{c}{Localization in environments with an additional object} \\
\midrule
R0 -- R1 & 3.2 $\pm$ 2.0 & 3.5 $\pm$ 2.1 & 8.6 $\pm$ 6.6 & 9.7 $\pm$ 7.3 \\
R1 -- R2 & 2.6 $\pm$ 1.8 & 3.0 $\pm$ 2.3 & 7.3 $\pm$ 5.3 & 9.0 $\pm$ 7.3 \\
R2 -- R3 & 3.2 $\pm$ 2.2 & 3.5 $\pm$ 2.2 & 7.0 $\pm$ 3.2 & 8.0 $\pm$ 4.1 \\
R3 -- R4 & 5.1 $\pm$ 2.1 & 5.3 $\pm$ 2.1 & 6.5 $\pm$ 4.3 & 7.6 $\pm$ 5.8 \\
\midrule
\multicolumn{5}{c}{Localization in environments with a removed object} \\
\midrule
R1 -- R0 & 1.7 $\pm$ 0.6 & 1.9 $\pm$ 0.7 & 4.3 $\pm$ 1.5 & 4.9 $\pm$ 2.1 \\
R2 -- R1 & 2.0 $\pm$ 1.0 & 2.4 $\pm$ 1.4 & 6.4 $\pm$ 4.4 & 7.8 $\pm$ 6.1 \\
R3 -- R2 & 2.0 $\pm$ 1.0 & 2.4 $\pm$ 1.3 & 6.4 $\pm$ 3.8 & 7.9 $\pm$ 5.4 \\
R4 -- R3 & 3.6 $\pm$ 2.6 & 4.0 $\pm$ 2.9 & 14.6 $\pm$ 4.0 & 16.8 $\pm$ 5.3 \\
\bottomrule
\end{tabular}\label{tab:real_results}
\end{table}

\begin{table}[t]
\centering
\caption{Localization error of individual motion capture rings.
Results pooled across all experiment conditions.}
\begin{tabular}{c|c|c}
\toprule
Ring & Translation MAE (cm) & Rotation MAE (deg) \\
\midrule
Top & 1.7 $\pm$ 1.3 & 4.8 $\pm$ 4.1  \\
Middle & 2.3 $\pm$ 1.2 & 7.0 $\pm$ 3.4 \\
Bottom (Tip) & 3.6 $\pm$ 1.9 & 9.9 $\pm$ 4.9 \\
\bottomrule
\end{tabular}\label{tab:ring_results}
\end{table}

\begin{figure}[t]
    \centering
    \includegraphics[width=\linewidth]{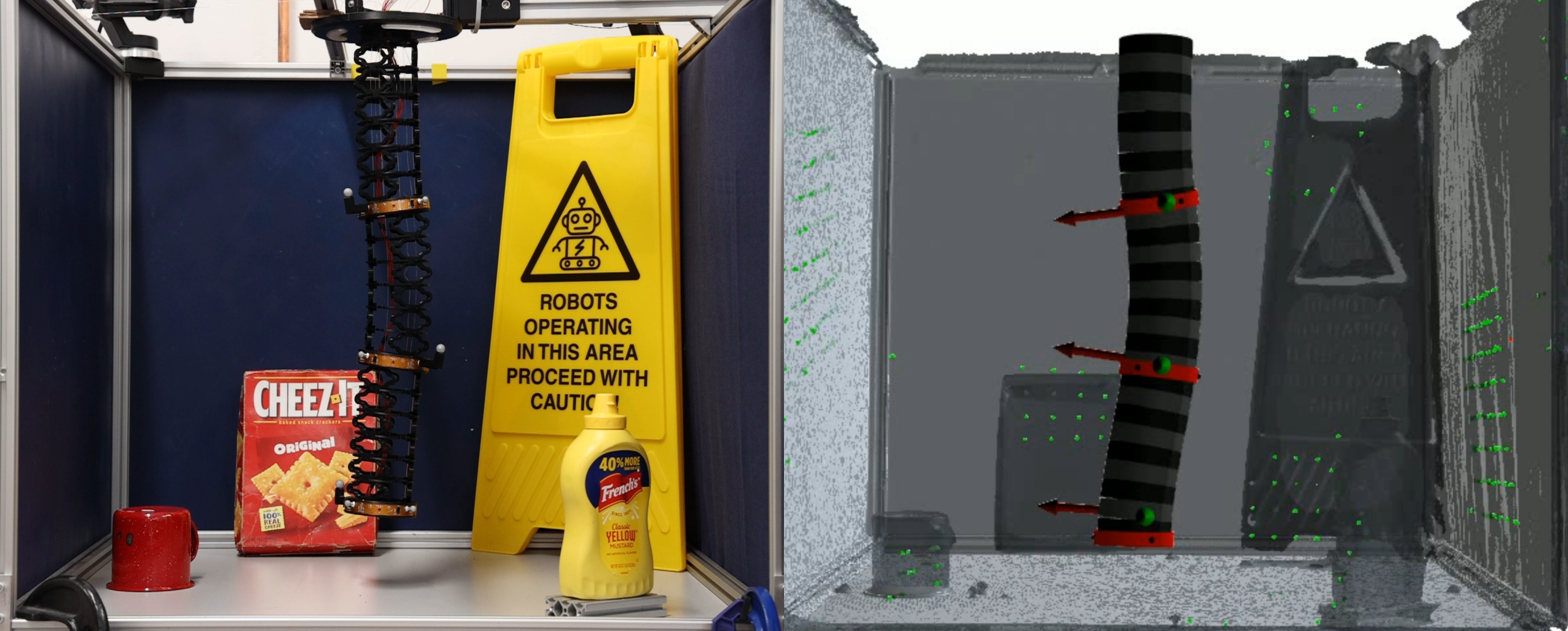}
    \includegraphics[width=\linewidth]{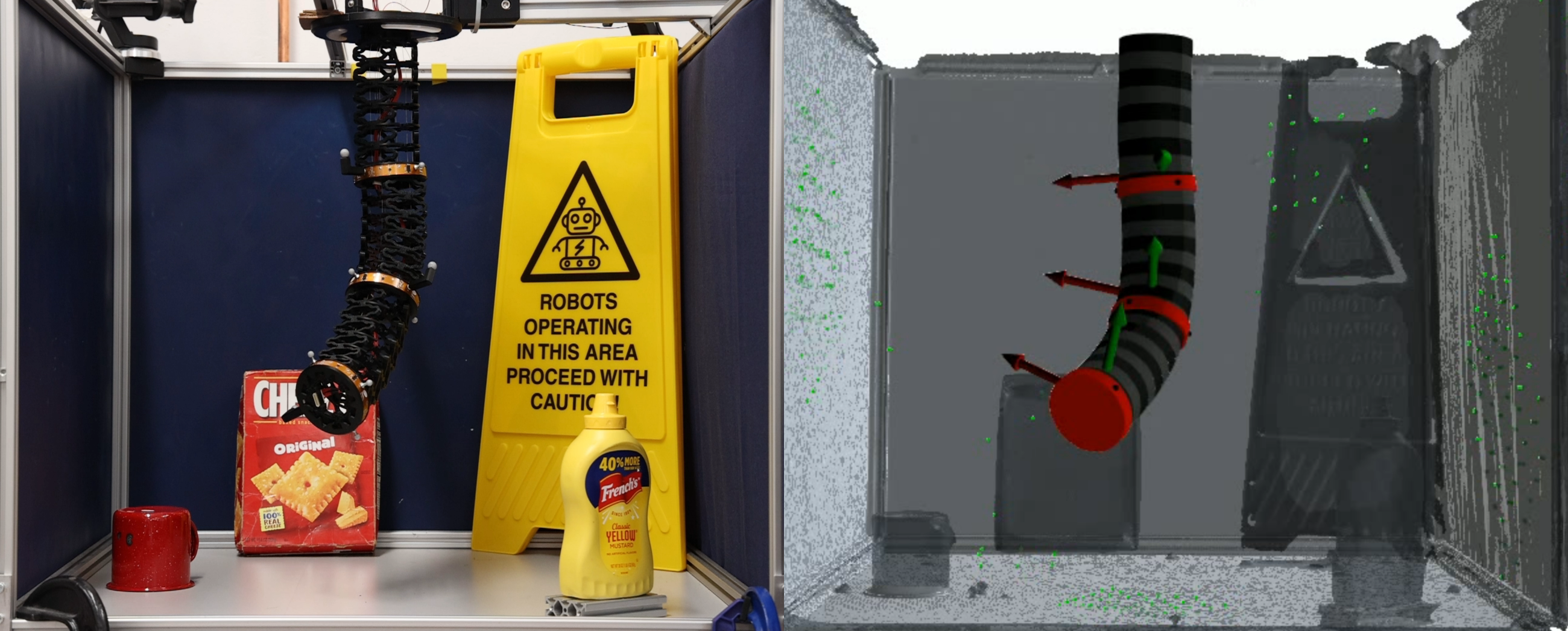}
    \includegraphics[width=\linewidth]{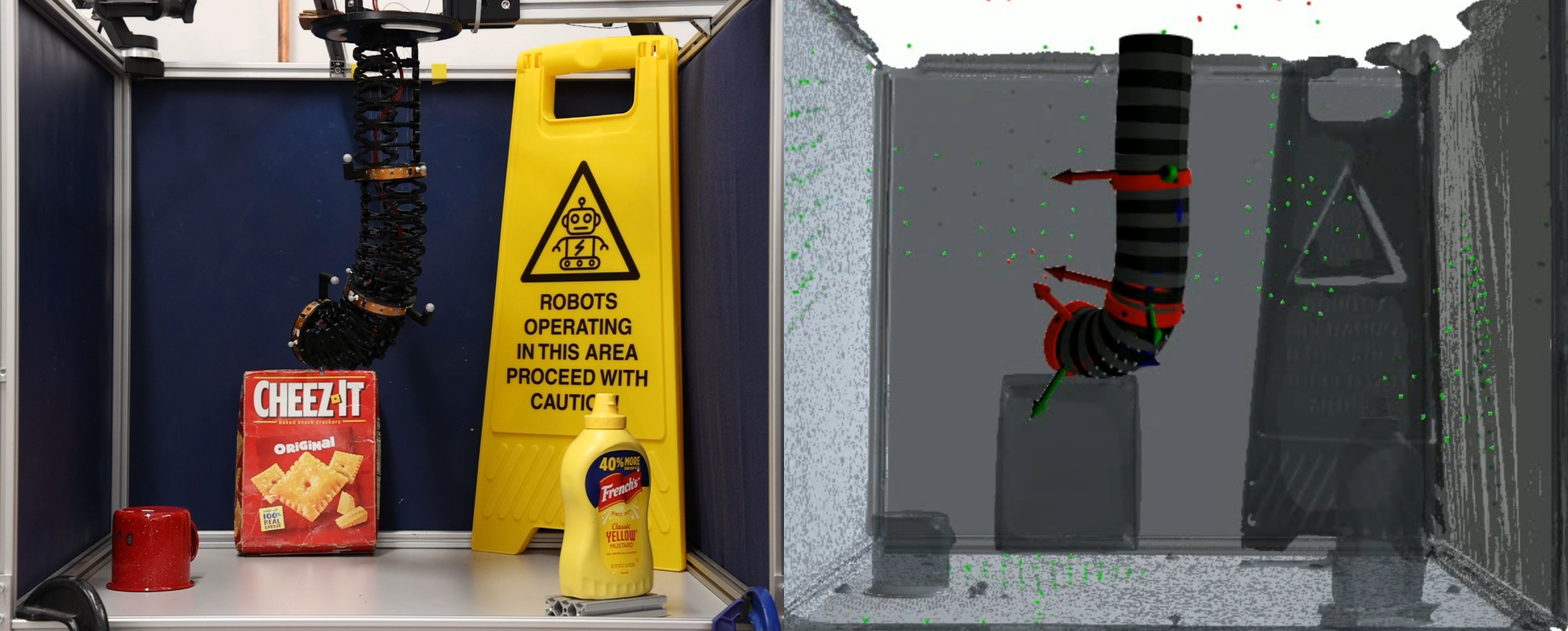}
    \caption{Side-by-side comparison of the real environment (left) and the estimated robot (right) for the median-performing trial (median average localization error). The estimated robot shape and trajectory reasonably match the ground truth, drawn as a coordinate axes, with an average position error at the marker locations of \SI{1.7}{cm} and rotational error of \ang{6.7}.}
    \label{fig:med_side_by_sides}
\end{figure}

\begin{figure}[t]
    \centering
    \includegraphics[width=\linewidth]{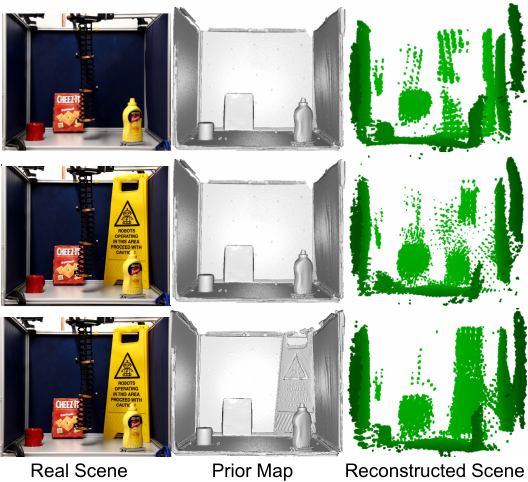}
    \caption{Real scenes (left), prior maps (middle), and the reconstructed scenes (right) via ToF measurements for the R3~--~R3 (top), R3~--~R4 (middle), and R4~--~R4 scenarios. A depth-based color gradient is applied to the point clouds for visual clarity. Objects in the scene with high reflectance or low incidence angle to the ToF sensors can lead to noisy measurements that affect reconstruction quality. This can be seen most clearly surrounding the cup on the left side of both scenes.}
    \label{fig:real_R4_pcd}
\end{figure}

Localization performance is evaluated for every combination of prior map and real environment (that varies by one object) as described in Section~\ref{sec:data_collection}. The results are summarized in Table~\ref{tab:real_results}. For trials where the prior map matches the real environment, the estimator performs accurately and consistently, achieving average position errors between \qtyrange{1.2}{2.5}{cm} and rotational ones between \qtyrange{3.7}{10.3}{\degree}.

When we introduce deviation between the true scenes and the prior, localization performance typically degrades. When an object is added to the environment, the localization error increases by an average of \SI{1.7}{cm} and \ang{1.0} across all trials. When an object is removed from the environment, the localization error increases by an average of \SI{0.5}{cm} and \ang{1.5}. The largest degradation in performance occurs when the large wet-floor sign is added to the environment, causing the average position error to increase to \SI{5.1}{cm} and rotational error to \ang{6.5} when added and to \SI{3.6}{cm} in position and \ang{14.6} in rotation when it is removed. 

The trial with the median performance (of all 61 combinations tested) is highlighted in Fig.~\ref{fig:med_side_by_sides}, showing a side-by-side comparison of the real robot and the estimated robot shape at several timesteps. The estimator clearly tracks the robot shape well, with this particular trial achieving an average position error of \SI{1.7}{cm} and rotational error of \ang{6.7} at the marker locations.

As with most continuum robot applications, the tip ring tends to have the worst localization performance due to error accumulation along the length of the robot. Average localization errors for each ring across all trials are summarized in Table~\ref{tab:ring_results}. While the ToF readings directly constrain the tip position along some axes, the distance from the strong prior information at the base of the robot leads to the tip disc being most susceptible to local minima. 

A side-by-side comparison of the real environment, prior map, and reconstructed scene via ToF measurements for the R4~--~R4, R3~--~R4, and R3~--~R3 trials is shown in Fig.~\ref{fig:real_R4_pcd}. In the cases where the prior map is accurate to the scene, reconstruction closely matches the real environment (per-point RMSE of \SI{0.8}{cm} for R4~--~R4, \SI{1.2}{cm} for R3~--~R3). When large anomalies degrade the localization performance, the reconstructed scene also deviates significantly from the real environment (per-point RMSE of \SI{1.4}{cm}). This indicates that improving localization performance in the presence of anomalies will also improve scene reconstruction accuracy.

\subsection{Limitations and Future Work}
\subsubsection{Sensor Degeneracies and Local Minima}

\begin{figure}[t]
    \centering
    \includegraphics[width=\linewidth]{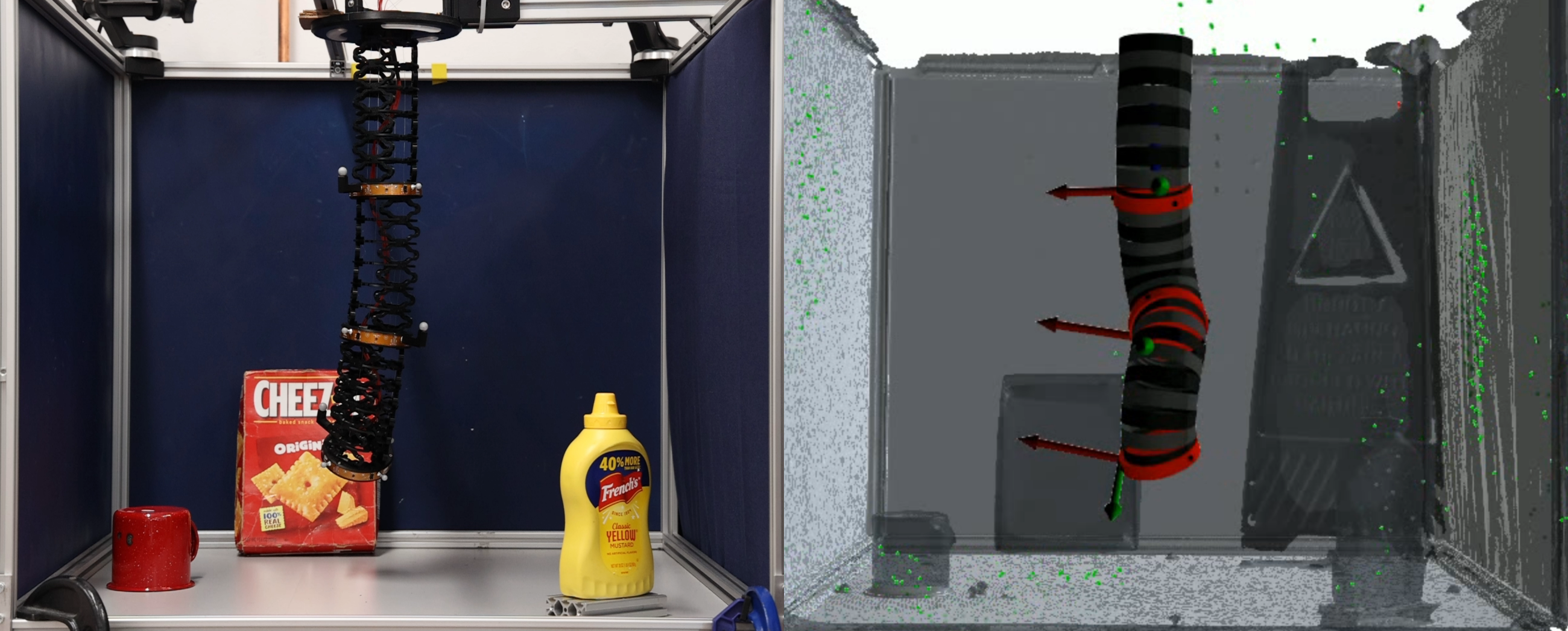}
    \includegraphics[width=\linewidth]{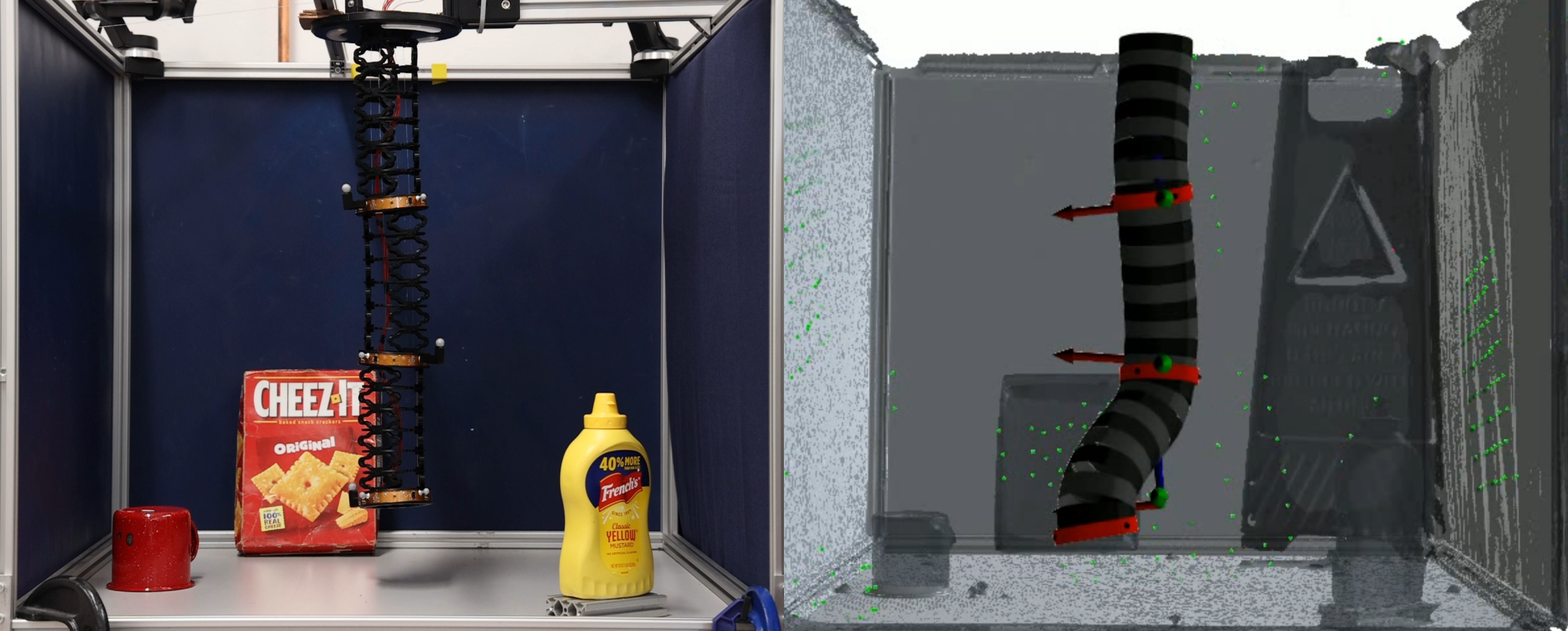}
    \includegraphics[width=\linewidth]{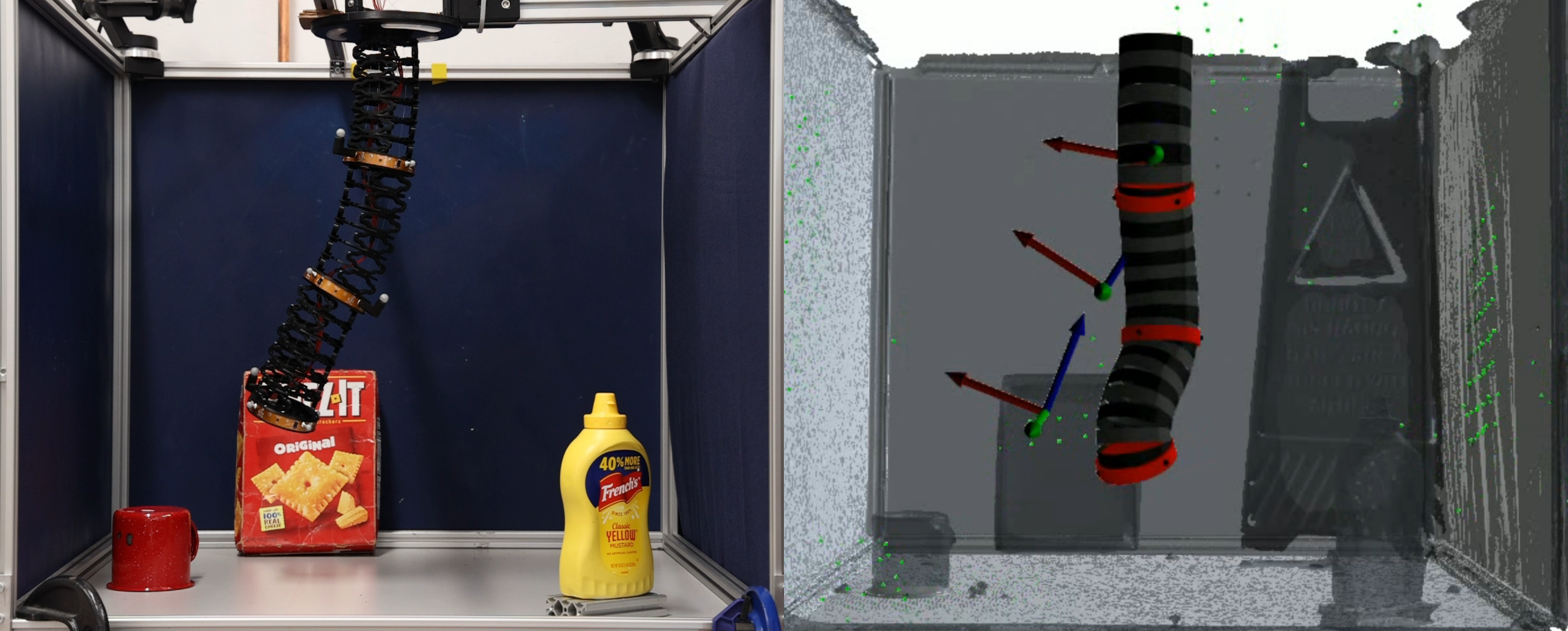}
    \caption{Example side-by-side comparison of the real environment (left) and the estimated robot (right) for failure cases in each of the three worst-performing trials (highest average localization error). In each case, the estimated robot shape and trajectory deviate significantly from the ground truth, getting stuck in local minima. Each of these scenarios features a map that is particularly challenging due to its deviation from the real system, where the large wet-floor sign has been removed.}
    \label{fig:worst_side_by_sides}
\end{figure}
As the ToF sensors acquire sparse measurements of the environment, even in feature-rich scenes, they are more susceptible to causing the estimator to get stuck in local minima. Without a method for detecting and escaping these minima, the estimator can produce large localization errors in certain scenarios. Example failure cases are shown in Fig.~\ref{fig:worst_side_by_sides}. \edit{A poor initialization, significant departure of the real scene from the map, and sudden movements can all lead to the estimator more easily converging to local minimum solutions.} This is not too surprising given the information being generated by the real-world robot. The experiment setup used in this work is particularly challenging with respect to degenerate scenarios due to the robot being extensible, the sensors having a limited field of view, and the scene having many symmetries. As demonstrated in the simulation result, the addition of more diverse sensor modalities (e.g.,~strain) helps alleviate this issue. \edit{The strain sensor information provides a restoring force effect to solutions that would otherwise have jumped to a local minimum.} The ISM330BX used as a gyroscope in this work also provides accelerometer measurements that could be incorporated to aid in this regard, though this is left to future work. 

\subsubsection{Estimator Smoothing and Tradeoffs}
The estimator framework used in this work introduces both temporal and spatial smoothing of the estimate. This effect is a double-edged sword. On one hand, it helps regularize the estimate and avoid overfitting to noisy measurements. On the other hand, it can prevent the estimator from accurately capturing rapid changes in the robot's shape over time. The local minima observed in the failure cases appear to be a result of both a highly deviated scene from the prior map and a sudden change in robot shape. 

\subsubsection{Prior Map Dependency}
As shown clearly in the results, the localization accuracy of this method is currently tied to the accuracy of the prior map. The larger this deviation, the worse the localization performance. While this is expected, it does limit the applicability of this method to scenarios where a reasonably accurate prior map can be obtained. Future work could explore methods for relaxing this requirement, incorporating live map updates, or otherwise improving robustness to map inaccuracies.

\subsubsection{Application: Anomaly Detection}
\begin{figure}[t]
    \centering
    \includegraphics[width=\linewidth]{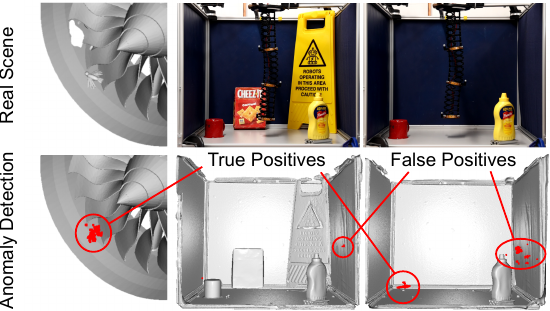}
    \caption{Real scene (top) with detected anomalies shown over prior map (bottom) highlighted in red. The detection is more successful in the simulated environment, where localization tends to be more accurate. Real-world experiments show false-positive detections near scene objects that corrode sensor readings. Further improvements to noise reduction and localization accuracy are required to make this a viable anomaly detection method. }
    \label{fig:anomaly_detection}
\end{figure}

One possible application of this work is anomaly detection in support of inspection tasks. To detect anomalies, each point in the reconstructed point cloud is compared against the prior map. A point is classified as an anomaly if its squared Mahalanobis distance to the nearest neighbor in the prior map exceeds a threshold $\tau$, i.e.
\begin{equation}
    (\bar{\boldsymbol{p}}_j - \boldsymbol{p}_{\text{nn}})^T (\boldsymbol{\Sigma}_{p_j} + \boldsymbol{\Sigma}_{nn})^{-1} (\bar{\boldsymbol{p}}_j - \boldsymbol{p}_{\text{nn}}) > \tau,
\end{equation} where $\boldsymbol{p}_{\text{nn}}$ is the nearest neighbor point in the prior map with prior covariance of $\boldsymbol{\Sigma}_{nn}$. Fig.~\ref{fig:anomaly_detection} highlights some results of this detector using the data from this work. With the simulated scenes performing slightly better with this rejection technique, additional work is required to improve the localization to the point of being able to reliably detect the changes in the real-world experiments.

\section{Conclusion} 
\label{sec:conclusion}
In this work, we show that full-body localization of continuum robots is achievable using sparse, low-resolution ToF sensors distributed along the robot body. Our proposed state estimation framework achieves an average localization accuracy of \SI{2.5}{cm} on an extensible continuum robot in 3D space, despite frequent sensing degeneracies and moderate deviations between the prior map and the true scene. These results demonstrate that compact, body-mounted depth sensors can support practical continuum robot localization without external tracking, while highlighting remaining challenges related to prior map dependence and local minima that motivate future work.

\section*{Acknowledgments}
\edit{We acknowledge the support of the Natural Sciences and Engineering Research Council of Canada (NSERC) [funding reference number RGPIN-2025-04343]. This research was also supported by the Canada Foundation for Innovation (CFI) through the John R. Evans Leaders Fund (JELF) [project number 40110].} This work was also supported by SESTOSENSO (HORIZON EUROPE Research and Innovation Actions) under GA number 101070310 and by the Engineering and Physical Sciences Research Council (EPSRC) Grant EP/V000748/1. We would like to thank Chengnan (Jimmy) Shentu and Nicholas Baldassini for their guidance in constructing environments and continuum robots in MuJoCo.

%% Use plainnat to work nicely with natbib. 

% \clearpage
\bibliographystyle{plainnat}
\bibliography{references}

\begin{thebibliography}{40}
\providecommand{\natexlab}[1]{#1}
\providecommand{\url}[1]{\texttt{#1}}
\expandafter\ifx\csname urlstyle\endcsname\relax
  \providecommand{\doi}[1]{doi: #1}\else
  \providecommand{\doi}{doi: \begingroup \urlstyle{rm}\Url}\fi

\bibitem[Abah et~al.(2022)Abah, Orekhov, Johnston, and Simaan]{Abah2022}
Colette Abah, Andrew~L. Orekhov, Garrison L.~H. Johnston, and Nabil Simaan.
\newblock A multi-modal sensor array for human–robot interaction and confined
  spaces exploration using continuum robots.
\newblock \emph{IEEE Sensors Journal}, 22\penalty0 (4):\penalty0 3585--3594,
  2022.
\newblock \doi{10.1109/JSEN.2021.3140002}.

\bibitem[Aerospace()]{GEAero}
GE~Aerospace.
\newblock Blade inspection toolkit.
\newblock URL
  \url{https://www.geaerospace.com/commercial/services/engine-maintenance-technologies/blade-inspection-tool}.

\bibitem[Barfoot(2017)]{Barfoot2017}
T~D Barfoot.
\newblock \emph{{State Estimation for Robotics}}.
\newblock Cambridge University Press, 2017.
\newblock ISBN 9781107159396.
\newblock \doi{10.1017/9781316671528}.

\bibitem[Borgstadt et~al.(2015)Borgstadt, Zinn, and Ferrier]{Borgstadt2015}
Justin~A. Borgstadt, Michael~R. Zinn, and Nicola~J. Ferrier.
\newblock Multi-modal localization algorithm for catheter interventions.
\newblock In \emph{2015 IEEE International Conference on Robotics and
  Automation (ICRA)}, pages 5350--5357, 2015.
\newblock \doi{10.1109/ICRA.2015.7139946}.

\bibitem[Burgner-Kahrs et~al.(2015)Burgner-Kahrs, Rucker, and
  Choset]{BurgnerKahrs2015}
Jessica Burgner-Kahrs, D.~Caleb Rucker, and Howie Choset.
\newblock Continuum robots for medical applications: A survey.
\newblock \emph{IEEE Transactions on Robotics}, 31\penalty0 (6):\penalty0
  1261--1280, 2015.
\newblock \doi{10.1109/TRO.2015.2489500}.

\bibitem[Burnett et~al.(2025)Burnett, Schoellig, and Barfoot]{Burnett2025}
Keenan Burnett, Angela~P. Schoellig, and Timothy~D. Barfoot.
\newblock Continuous-time radar-inertial and lidar-inertial odometry using a
  gaussian process motion prior.
\newblock \emph{IEEE Transactions on Robotics}, 41:\penalty0 1059--1076, 2025.
\newblock \doi{10.1109/TRO.2024.3521856}.

\bibitem[Caroleo et~al.()Caroleo, Albini, and Maiolino]{Caroleo2025b}
Giammarco Caroleo, Alessandro Albini, and Perla Maiolino.
\newblock Soft robot localization using distributed miniaturized time-of-flight
  sensors.
\newblock In \emph{2025 {IEEE} 8th International Conference on Soft Robotics
  ({RoboSoft})}, pages 1--6.
\newblock \doi{10.1109/RoboSoft63089.2025.11020858}.
\newblock {ISSN}: 2769-4534.

\bibitem[Caroleo et~al.(2025)Caroleo, Albini, De~Martini, Barfoot, and
  Maiolino]{Caroleo2025a}
Giammarco Caroleo, Alessandro Albini, Daniele De~Martini, Timothy~D. Barfoot,
  and Perla Maiolino.
\newblock Tiny lidars for manipulator self-awareness: Sensor characterization
  and initial localization experiments.
\newblock In \emph{2025 IEEE/RSJ International Conference on Intelligent Robots
  and Systems (IROS)}, pages 12940--12946, 2025.
\newblock \doi{10.1109/IROS60139.2025.11246575}.

\bibitem[{CATIAV5FTW}(2025)]{Catiav5ftw2025}
{CATIAV5FTW}.
\newblock “3d printable jet engine” [3d model].
\newblock
  \url{https://www.printables.com/model/572421-3d-printable-jet-engine}, 2025.
\newblock Adapted by the authors. Original work licensed under Creative Commons
  Attribution-NonCommercial 4.0 International (CC BY-NC 4.0).

\bibitem[Chang et~al.(2025)Chang, Dai, Wang, Axinte, and Dong]{Chang2025}
Jung-Che Chang, Hengtai Dai, Xi~Wang, Dragos Axinte, and Xin Dong.
\newblock Development of a continuum robot with inflatable stiffness-adjustable
  elements for in-situ repair of aeroengines.
\newblock \emph{Robotics and Computer-Integrated Manufacturing}, 95:\penalty0
  103018, 2025.
\newblock ISSN 0736-5845.
\newblock \doi{https://doi.org/10.1016/j.rcim.2025.103018}.

\bibitem[Demantk\'e et~al.(2011)Demantk\'e, Mallet, David, and
  Vallet]{Demantke2011}
J.~Demantk\'e, C.~Mallet, N.~David, and B.~Vallet.
\newblock Dimensionality based scale selection in 3d lidar point clouds.
\newblock \emph{The International Archives of the Photogrammetry, Remote
  Sensing and Spatial Information Sciences}, XXXVIII-5/W12:\penalty0 97--102,
  2011.

\bibitem[Dong et~al.(2017)Dong, Axinte, Palmer, Cobos, Raffles, Rabani, and
  Kell]{Dong2017}
X.~Dong, D.~Axinte, D.~Palmer, S.~Cobos, M.~Raffles, A.~Rabani, and J.~Kell.
\newblock Development of a slender continuum robotic system for on-wing
  inspection/repair of gas turbine engines.
\newblock \emph{Robotics and Computer-Integrated Manufacturing}, 44:\penalty0
  218--229, 2017.
\newblock ISSN 0736-5845.
\newblock \doi{https://doi.org/10.1016/j.rcim.2016.09.004}.

\bibitem[Dupont et~al.(2022)Dupont, Simaan, Choset, and Rucker]{Dupont2022}
Pierre~E. Dupont, Nabil Simaan, Howie Choset, and Caleb Rucker.
\newblock Continuum robots for medical interventions.
\newblock \emph{Proceedings of the IEEE}, 110\penalty0 (7):\penalty0 847--870,
  2022.
\newblock \doi{10.1109/JPROC.2022.3141338}.

\bibitem[Ferguson et~al.(2024)Ferguson, Rucker, and Webster]{Ferguson2024}
James~M. Ferguson, D.~Caleb Rucker, and Robert~J. Webster.
\newblock Unified shape and external load state estimation for continuum
  robots.
\newblock \emph{IEEE Transactions on Robotics}, 40:\penalty0 1813--1827, 2024.
\newblock \doi{10.1109/TRO.2024.3360950}.

\bibitem[Franz et~al.(2014)Franz, Haidegger, Birkfellner, Cleary, Peters, and
  Maier-Hein]{Franz2014}
Alfred~M. Franz, Tamás Haidegger, Wolfgang Birkfellner, Kevin Cleary, Terry~M.
  Peters, and Lena Maier-Hein.
\newblock Electromagnetic tracking in medicine—a review of technology,
  validation, and applications.
\newblock \emph{IEEE Transactions on Medical Imaging}, 33\penalty0
  (8):\penalty0 1702--1725, 2014.
\newblock \doi{10.1109/TMI.2014.2321777}.

\bibitem[Girerd et~al.(2020)Girerd, Kudryavtsev, Rougeot, Renaud, Rabenorosoa,
  and Tamadazte]{Girerd2020}
Cedric Girerd, Andrey~V. Kudryavtsev, Patrick Rougeot, Pierre Renaud, Kanty
  Rabenorosoa, and Brahim Tamadazte.
\newblock Slam-based follow-the-leader deployment of concentric tube robots.
\newblock \emph{IEEE Robotics and Automation Letters}, 5\penalty0 (2):\penalty0
  548--555, 2020.
\newblock \doi{10.1109/LRA.2019.2963821}.

\bibitem[Holland and Welsch(1977)]{Holland1977}
Paul~W. Holland and Roy~E. Welsch.
\newblock Robust regression using iteratively reweighted least-squares.
\newblock \emph{Communications in Statistics - Theory and Methods}, 6\penalty0
  (9):\penalty0 813--827, 1977.
\newblock \doi{10.1080/03610927708827533}.

\bibitem[Iwao et~al.(2025)Iwao, Arita, and Tahara]{Iwao2025}
Kengo Iwao, Hikaru Arita, and Kenji Tahara.
\newblock State estimation and environment recognition for articulated
  structures via proximity sensors distributed over the whole body.
\newblock \emph{IEEE Robotics and Automation Letters}, 10\penalty0
  (3):\penalty0 3030--3037, 2025.
\newblock \doi{10.1109/LRA.2025.3539117}.

\bibitem[Kim et~al.(2014)Kim, Ha, Park, and Dupont]{Kim2014}
Beobkyoon Kim, Junhyoung Ha, Frank~C. Park, and Pierre~E. Dupont.
\newblock Optimizing curvature sensor placement for fast, accurate shape
  sensing of continuum robots.
\newblock In \emph{2014 IEEE International Conference on Robotics and
  Automation (ICRA)}, pages 5374--5379, 2014.
\newblock \doi{10.1109/ICRA.2014.6907649}.

\bibitem[Koolwal et~al.(2010)Koolwal, Barbagli, Carlson, and
  Liang]{Koolwal2010}
Aditya~Brij Koolwal, Federico Barbagli, Christopher Carlson, and David Liang.
\newblock An ultrasound-based localization algorithm for catheter ablation
  guidance in the left atrium.
\newblock \emph{The International Journal of Robotics Research}, 29\penalty0
  (6):\penalty0 643--665, 2010.
\newblock \doi{10.1177/0278364909105332}.

\bibitem[Lilge et~al.(2022)Lilge, Barfoot, and Burgner-Kahrs]{Lilge2022}
Sven Lilge, Timothy~D. Barfoot, and Jessica Burgner-Kahrs.
\newblock Continuum robot state estimation using gaussian process regression on
  se(3).
\newblock \emph{The International Journal of Robotics Research}, 41\penalty0
  (13-14):\penalty0 1099--1120, 2022.
\newblock \doi{10.1177/02783649221128843}.

\bibitem[Mahoney et~al.()Mahoney, Bruns, Swaney, and Webster]{Mahoney2016}
Arthur~W. Mahoney, Trevor~L. Bruns, Philip~J. Swaney, and Robert~J. Webster.
\newblock On the inseparable nature of sensor selection, sensor placement, and
  state estimation for continuum robots or “where to put your sensors and how
  to use them”.
\newblock In \emph{2016 {IEEE} International Conference on Robotics and
  Automation ({ICRA})}, pages 4472--4478.
\newblock \doi{10.1109/ICRA.2016.7487646}.

\bibitem[Rao et~al.(2021)Rao, Peyron, Lilge, and Burgner-Kahrs]{Rao2021}
Priyanka Rao, Quentin Peyron, Sven Lilge, and Jessica Burgner-Kahrs.
\newblock How to model {Tendon-Driven} continuum robots and benchmark modelling
  performance.
\newblock \emph{Front Robot AI}, 7:\penalty0 630245, February 2021.

\bibitem[Reiter et~al.(2011)Reiter, Goldman, Bajo, Iliopoulos, Simaan, and
  Allen]{Reiter2011}
Austin Reiter, Roger~E. Goldman, Andrea Bajo, Konstantinos Iliopoulos, Nabil
  Simaan, and Peter~K. Allen.
\newblock A learning algorithm for visual pose estimation of continuum robots.
\newblock In \emph{2011 IEEE/RSJ International Conference on Intelligent Robots
  and Systems}, pages 2390--2396, 2011.
\newblock \doi{10.1109/IROS.2011.6094947}.

\bibitem[Rone and Ben-Tzvi(2013)]{Rone2013}
William~S. Rone and Pinhas Ben-Tzvi.
\newblock Multi-segment continuum robot shape estimation using passive cable
  displacement.
\newblock In \emph{2013 IEEE International Symposium on Robotic and Sensors
  Environments (ROSE)}, pages 37--42, 2013.
\newblock \doi{10.1109/ROSE.2013.6698415}.

\bibitem[Russo et~al.(2023)Russo, Sadati, Dong, Mohammad, Walker, Bergeles, Xu,
  and Axinte]{Russo2023}
Matteo Russo, Seyed Mohammad~Hadi Sadati, Xin Dong, Abdelkhalick Mohammad,
  Ian~D. Walker, Christos Bergeles, Kai Xu, and Dragos~A. Axinte.
\newblock Continuum robots: An overview.
\newblock \emph{Advanced Intelligent Systems}, 5\penalty0 (5):\penalty0
  2200367, 2023.
\newblock \doi{https://doi.org/10.1002/aisy.202200367}.

\bibitem[Shentu et~al.(2023)Shentu, Li, Chen, Dewi, Lindell, and
  Burgner-Kahrs]{Shentu2023}
Chengnan Shentu, Enxu Li, Chaojun Chen, Puspita~T Dewi, David~B Lindell, and
  Jessica Burgner-Kahrs.
\newblock Moss: Monocular shape sensing for continuum robots.
\newblock \emph{IEEE Robotics and Automation Letters}, 9\penalty0 (2):\penalty0
  1524--1531, 2023.

\bibitem[Shi et~al.(2017)Shi, Luo, Qi, Li, Song, Najdovski, Fukuda, and
  Ren]{Shi2017}
Chaoyang Shi, Xiongbiao Luo, Peng Qi, Tianliang Li, Shuang Song, Zoran
  Najdovski, Toshio Fukuda, and Hongliang Ren.
\newblock Shape sensing techniques for continuum robots in minimally invasive
  surgery: A survey.
\newblock \emph{IEEE Transactions on Biomedical Engineering}, 64\penalty0
  (8):\penalty0 1665--1678, 2017.
\newblock \doi{10.1109/TBME.2016.2622361}.

\bibitem[Sincak et~al.(2024)Sincak, Prada, Miková, Mykhailyshyn, Varga, Merva,
  and Virgala]{Sincak2024}
Peter~Jan Sincak, Erik Prada, Ľubica Miková, Roman Mykhailyshyn, Martin
  Varga, Tomas Merva, and Ivan Virgala.
\newblock Sensing of continuum robots: A review.
\newblock \emph{Sensors}, 24\penalty0 (4), 2024.
\newblock ISSN 1424-8220.
\newblock \doi{10.3390/s24041311}.

\bibitem[Sorensen et~al.(2021)Sorensen, Hyatt, Ricks, Nielsen, and
  Killpack]{Sorensen2021}
Christian Sorensen, Phillip Hyatt, Matthew Ricks, Seth Nielsen, and Marc~D.
  Killpack.
\newblock Soft robot configuration estimation and control using simultaneous
  localization and mapping.
\newblock In \emph{2021 IEEE/RSJ International Conference on Intelligent Robots
  and Systems (IROS)}, pages 616--623, 2021.
\newblock \doi{10.1109/IROS51168.2021.9635896}.

\bibitem[Store(2026)]{3d_scanner}
Creality Store.
\newblock Cr-scan raptor 3d scanner, 2026.
\newblock URL
  \url{https://store.creality.com/uk/products/cr-scan-raptor-3d-scanner}.

\bibitem[Teetaert et~al.(2025)Teetaert, Lilge, Burgner-Kahrs, and
  Barfoot]{Teetaert2025a}
Spencer Teetaert, Sven Lilge, Jessica Burgner-Kahrs, and Timothy~D. Barfoot.
\newblock A stochastic framework for continuous-time state estimation of
  continuum robots.
\newblock 2025.
\newblock {\it \href{https://arxiv.org/abs/2510.01381}{arXiv:2510.01381}}.

\bibitem[Teetaert et~al.(2026)Teetaert, Lilge, Burgner-Kahrs, and
  Barfoot]{Teetaert2025b}
Spencer Teetaert, Sven Lilge, Jessica Burgner-Kahrs, and Timothy~D. Barfoot.
\newblock A sliding-window filter for online continuous-time continuum robot
  state estimation.
\newblock In \emph{2026 IEEE 9th International Conference on Soft Robotics
  (RoboSoft)}, pages 239--246, 2026.

\bibitem[Todorov et~al.(2012)Todorov, Erez, and Tassa]{Todorov2012}
Emanuel Todorov, Tom Erez, and Yuval Tassa.
\newblock Mujoco: A physics engine for model-based control.
\newblock In \emph{2012 IEEE/RSJ International Conference on Intelligent Robots
  and Systems}, pages 5026--5033. IEEE, 2012.
\newblock \doi{10.1109/IROS.2012.6386109}.

\bibitem[Vaquero et~al.(2024)Vaquero, Daddi, Thakker, Paton, Jasour, Strub,
  Swan, Royce, Gildner, Tosi, Veismann, Gavrilov, Marteau, Bowkett,
  de~Mola~Lemus, Nakka, Hockman, Orekhov, Hasseler, Leake, Nuernberger,
  Proença, Reid, Talbot, Georgiev, Pailevanian, Archanian, Ambrose, Jasper,
  Etheredge, Roman, Levine, Otsu, Yearicks, Melikyan, Rieber, Carpenter, Nash,
  Jain, Shiraishi, Robinson, Travers, Choset, Burdick, Gardner, Cable, Ingham,
  and Ono]{Vaquero2024}
T.~S. Vaquero, G.~Daddi, R.~Thakker, M.~Paton, A.~Jasour, M.~P. Strub, R.~M.
  Swan, R.~Royce, M.~Gildner, P.~Tosi, M.~Veismann, P.~Gavrilov, E.~Marteau,
  J.~Bowkett, D.~Loret de~Mola~Lemus, Y.~Nakka, B.~Hockman, A.~Orekhov, T.~D.
  Hasseler, C.~Leake, B.~Nuernberger, P.~Proença, W.~Reid, W.~Talbot,
  N.~Georgiev, T.~Pailevanian, A.~Archanian, E.~Ambrose, J.~Jasper,
  R.~Etheredge, C.~Roman, D.~Levine, K.~Otsu, S.~Yearicks, H.~Melikyan, R.~R.
  Rieber, K.~Carpenter, J.~Nash, A.~Jain, L.~Shiraishi, M.~Robinson,
  M.~Travers, H.~Choset, J.~Burdick, A.~Gardner, M.~Cable, M.~Ingham, and
  M.~Ono.
\newblock Eels: Autonomous snake-like robot with task and motion planning
  capabilities for ice world exploration.
\newblock \emph{Science Robotics}, 9\penalty0 (88):\penalty0 eadh8332, 2024.
\newblock \doi{10.1126/scirobotics.adh8332}.

\bibitem[Wang et~al.(2023)Wang, Dai, Tong, Wang, Fang, Xie, Liu, Au, and
  Kwok]{Wang2023}
Xiaomei Wang, Jing Dai, Hon-Sing Tong, Kui Wang, Ge~Fang, Xiaochen Xie, Yun-Hui
  Liu, Kwok Wai~Samuel Au, and Ka-Wai Kwok.
\newblock Learning-based visual-strain fusion for eye-in-hand continuum robot
  pose estimation and control.
\newblock \emph{IEEE Transactions on Robotics}, 39\penalty0 (3):\penalty0
  2448--2467, 2023.
\newblock \doi{10.1109/TRO.2023.3240556}.

\bibitem[Wang et~al.(2020)Wang, Ju, Yun, Yao, Wang, Guo, and Chen]{Wang2020}
Yaming Wang, Feng Ju, Yahui Yun, Jiafeng Yao, Yaoyao Wang, Hao Guo, and Bai
  Chen.
\newblock An inspection continuum robot with tactile sensor based on electrical
  impedance tomography for exploration and navigation in unknown environment.
\newblock \emph{The Industrial Robot}, 47\penalty0 (1):\penalty0 121--130,
  2020.

\bibitem[Wei et~al.(2022)Wei, Liu, Ling, and Su]{Wei2022}
Xinyue Wei, Minghua Liu, Zhan Ling, and Hao Su.
\newblock Approximate convex decomposition for 3d meshes with collision-aware
  concavity and tree search.
\newblock \emph{ACM Transactions on Graphics (TOG)}, 41\penalty0 (4):\penalty0
  1--18, 2022.

\bibitem[Yamauchi et~al.(2022)Yamauchi, Ambe, Nagano, Konyo, Bando, Ito,
  Arnold, Yamazaki, Itoyama, Okatani, Okuno, and Tadokoro]{Yamauchi2022}
Yu~Yamauchi, Yuichi Ambe, Hikaru Nagano, Masashi Konyo, Yoshiaki Bando, Eisuke
  Ito, Solvi Arnold, Kimitoshi Yamazaki, Katsutoshi Itoyama, Takayuki Okatani,
  Hiroshi~G Okuno, and Satoshi Tadokoro.
\newblock Development of a continuum robot enhanced with distributed sensors
  for search and rescue.
\newblock \emph{ROBOMECH J.}, 9\penalty0 (1), December 2022.

\bibitem[Zhang et~al.(2022)Zhang, Fang, Xiang, Sun, Xue, Jin, Qiu, Xiong, Wang,
  and Lu]{Zhang2022}
Jingyu Zhang, Qin Fang, Pingyu Xiang, Danying Sun, Yanan Xue, Rui Jin, Ke~Qiu,
  Rong Xiong, Yue Wang, and Haojian Lu.
\newblock A survey on design, actuation, modeling, and control of continuum
  robot.
\newblock \emph{Cyborg and Bionic Systems}, 2022:\penalty0 9754697, 2022.
\newblock \doi{10.34133/2022/9754697}.

\end{thebibliography}

\end{document}